\pdfoutput=1
\documentclass{agujournal}

\usepackage{graphics}
\usepackage{epstopdf}
\usepackage{longtable}
\usepackage{float}
\DeclareGraphicsExtensions{.png,.pdf}

\journalname{arXiv}

\begin{document}

\title{Prolongation of SMAP to Spatio-temporally Seamless Coverage of Continental US Using a Deep Learning Neural Network}

\authors{Kuai Fang\affil{1}, Chaopeng Shen\affil{1}, Daniel Kifer\affil{2}, Xiao Yang\affil{2}}

\affiliation{1}{Department of Civil and Environmental Engineering,Pennsylvania State University, University Park, Pennsylvania, USA.}
\affiliation{2}{Department of Computer Science and Engineering, Pennsylvania State University, University Park, Pennsylvania, USA.}

\correspondingauthor{Chaopeng Shen}{cshen@engr.psu.edu}

\begin{keypoints}
\item With 2 years of data, SMAP L3 data can be extended at high fidelity using a deep learning network (LSTM), showing potential for hindcasting.
\item Despite significant, spatially-varying bias in Land Surface Models, LSTM can remove bias, correct moisture climatology, and capture extremes.
\item LSTM is more generalizable than other tested simpler methods, and its strength seems to derive from its memory and ability to accommodate large data.

\end{keypoints}

\begin{abstract}
The Soil Moisture Active Passive (SMAP) mission has delivered valuable sensing of surface soil moisture since 2015. However, it has a short time span and irregular revisit schedule. Utilizing a state-of-the-art time-series deep learning neural network, Long Short-Term Memory (LSTM), we created a system that predicts SMAP level-3 soil moisture data with atmospheric forcing, model-simulated moisture, and static physiographic attributes as inputs. The system removes most of the bias with model simulations and improves predicted moisture climatology, achieving small test root-mean-squared error (<0.035) and high correlation coefficient >0.87 for over 75\% of Continental United States, including the forested Southeast. As the first application of LSTM in hydrology, we show the proposed network avoids overfitting and is robust for both temporal and spatial extrapolation tests. LSTM generalizes well across regions with distinct climates and physiography. With high fidelity to SMAP, LSTM shows great potential for hindcasting, data assimilation, and weather forecasting.
\end{abstract}

\section{Introduction} 


Soil moisture is a key variable that controls various hydrologic processes, including infiltration, evapotranspiration and subsurface flow. It is of central importance to drought monitoring \citep{Narasimhan2005}, floods prediction \citep{Norbiato2008}, weather forecasting \citep{Koster2004}, irrigation planning and many other scientifically- and socially-important applications. Launched in 2015, NASA's Soil Moisture Active Passive (SMAP) satellite mission \citep{Entekhabi2010} is designed to measure top 5 cm soil moisture globally with a standard deviation of $\pm0.04$ cm$^3$/cm$^3$ volumetric ratio when vegetation water content (VWC) $\leq5 $kg/m$^2$ \citep{ONeill2012}. It achieved this goal in most core evaluation sites \citep{Colliander2017,Jackson2016}. Notwithstanding its great value, SMAP passive radiometer-based observations only have a short time span (since April 2015) with an irregular revisit time of 2-3 days, which makes it difficult to observe soil moisture responses immediately after storms or snowmelt.



Compared to SMAP's limited resolution and time span, land surface models (LSMs), {\it e.g.}, VIC \citep{Nijssen2001}, Noah \citep{Ek2003}, CLSM \citep{Koster2000} and MOS \citep{Koster1994}, simulate soil moisture seamlessly over much longer time spans. Despite their frequent use, these models may differ significantly from observations \citep{Leeper2017, Yuan2017,Dirmeyer2016,Xia2015}. Biases (mean difference from the observed) are notable in all models evaluated in \cite{Xia2015}. Their error patterns generally vary by region, model, season, and soil depths, yet there are systematic patterns in them. For example, moisture tends to be over-estimated in the arid western CONUS and under-estimated in wetter eastern US \citep{Yuan2017}; the Noah model tends to under-estimate moisture in wet seasons and over-estimate in dry seasons \citep{Xia2015}. These systematic error patterns could be exploited to improve predictions. 

To correct systematic model errors, we turn to deep learning (DL), a rebranding of artificial neural network. DL has made revolutionary strides in recent years and helped to solve problems that have resisted artificial intelligence for decades \citep{LeCun2015}. With earlier-generation machine learning methods, human experts extract features from data that are strongly correlated with dependent variables. DL, on the other hand, automatically extracts abstract features through their hidden layers. Two highly successful network structures are convolutional neural networks (CNN) for image-domain tasks \citep{Krizhevsky2012}, and Long Short-Term Memory (LSTM) \citep{Hochreiter1997,Greff2015} for time-domain tasks, although the separation is not absolute. No study, to the best of the authors' knowledge, has employed time series deep learning in hydrology. Given DL's success in other scientific disciplines \citep{Voosen2017}, it is plausible that DL can capture model error patterns that humans have yet come to explicitly formulate. 



The parameter space of deep networks is substantially large in order to provide the flexibility in mapping diverse, complex functions. Thus one might be concerned about overfitting, which means coefficients are fitted to noise rather than meaningful information. However, there are recent breakthroughs in regularization techniques, e.g., Dropout \citep{Srivastava2014}, which penalize overfitting and reduce mutually dependent coefficients. Nevertheless, since LSTM has not been applied to hydrology, it is important to examine its robustness compared to conventional statistical methods.



The central hypothesis of this work is that with two years of SMAP data, LSTM can learn patterns in soil moisture dynamics and LSMs errors, and by utilizing them, can SMAP data over long time spans. Our objectives are: (1) to produce a seamless top-surface soil moisture dataset for continental United States (CONUS) with high fidelity to SMAP data; (2) to provide an initial investigation of LSTM's capability in correcting process-based model errors; (3) to compare LSTM's generalization capability to conventional methods in spatial and temporal extrapolation tests. Here, by "high fidelity", we mean a high consistency with the target data resulting in its faithful reproduction. SMAP's retrieval algorithm for the passive product derives soil moisture from brightness temperature readings using radiative transfer and soil dielectric models \citep{ONeill2012}, thus it also incurs biases \citep{Colliander2017}. Nevertheless, a high-fidelity hindcast product has a wide range of applications, e.g., data mining of past fire hazards, calibration of hydrologic models, or benchmarking satellite product with historical {\it{in-situ}} data.


\section{Methods and Datasets}
As an overview, we trained an LSTM network to predict SMAP L3 product with, as inputs, atmospheric forcing time series, LSM-simulated surface soil moisture and static physiographic attributes. We compared LSTM to regularized multiple linear regression, auto-regressive models, and a simple one-layer feedforward neural network. Their performances were tested by (i) temporal generalization test: training over one year and testing over another; (ii) regular spatial generalization test: training over a uniformly down-selected subset of SMAP pixels and testing over other cells; and (iii) regional holdout test: training on some sub-regions of CONUS and test on the rest. All data sources are aggregated to a daily time scale and interpolated to SMAP L3 grid. Each SMAP pixel is an {\it instance}.  

\subsection{Data sources and inputs}


For the learning target, we focus on the L3 passive radiometer product (L3\_SM\_P) which combines swaths available in each day. The spatial resolution of L3\_SM\_P is ~36 km. For inputs, we obtained atmospheric forcing data including precipitation, temperature, radiation, humidity and wind speed from North-American Land Data Assimilation System phase II (NLDAS-2) \citep{Xia2015}. NLDAS-2 also provides, from 1979 to present, simulations of land surface states and fluxes by several LSMs. We chose Noah's (and also compared with MOS's) outputs \citep{Ek2003} because it ranks in the middle among models \citep{Xia2015} and is not as extensively calibrated as some other models, e.g., SAC. Our work does not require the best LSM, as we can observe how LSTM and other methods correct LSM errors. Noah has 4 soil layers which are of depths 0-10, 10-40, 40-100 and 100-200 cm, respectively. To match with the 0-5 cm sensed by SMAP, we tested: (i) directly using 0-10 cm data; (ii) polynomial interpolation; and (iii) integral interpolation where we find polynomials whose integrals agree with Noah-simulated values.

Static physiographic attributes (Table S1 in SI) include sand, silt and clay percentages, bulk density and soil capacity from ISRIC-WISE \citep{Batjes1995}. County-level annual-average irrigation data for 2010 \citep{USGS2016} was overlaid with landuse data to assign irrigation in each county to agricultural land uses. The values are then aggregated to SMAP grid. Also among inputs are SMAP product flags that indicate mountainous terrain, land cover classes, VWC, urban area, water body fraction and data quality (time-averaged). SMAP product flags indicate lower data quality in dense vegetated or forest area. However, instead of removing all regions labeled as low-quality, we hypothesize that including the flags as inputs allows LSTM to implicitly assign less focus to high-uncertainty regions. 
\subsection{LSTM setup}
As a type of Recurrent Neural Network (RNN), LSTM makes use of sequential information by updating states based on both inputs of the current time step ($x^{t}$) and network states of previous time steps, as illustrated in Figure S1 in Supporting information (SI). Following the notation in \cite{Lipton2015}, we can write an LSTM as LSTM : $x^{(t)}, h^{(t-1)}, s^{(t-1)} \to h^{(t)}, s^{(t)}$. The update formula are:
\begin{eqnarray}
\label{eq_lstm}
\textrm{(input node)} &g^{(t)} &=\tanh(W_{gx} x^{(t)} + W_{gh} h^{(t-1)} + b_{g}) \\ 
\textrm{(input gate)} &i^{(t)} &=\sigma(W_{ix} x^{(t)} + W_{ih} h^{(t-1)} + b_{i}) \\ 
\textrm{(forget gate)} &f^{(t)} &=\sigma(W_{fx} x^{(t)} + W_{fh} h^{(t-1)} + b_{f}) \\ 
\textrm{(output gate)} &o^{(t)} &=\sigma(W_{ox} x^{(t)} + W_{oh} h^{(t-1)} + b_{o}) \\ 
\textrm{(cell state)} &s^{(t)} &=g^{(t)} \odot i^{(t)} + s^{(t-1)} \odot f^{(t)}   \\ 
\textrm{(hidden gate)} &h^{(t)} &=\tanh(s^{(t)}) \odot o^{(t)} \\
\textrm{(output layer)} &y^{(t)} & = W_{hy} h^{(t)}+b_y  
\end{eqnarray} 
%
where $\sigma$ is the sigmoidal function, $\odot$ is element-wise multiplication, $x^{(t)}$ is the input vector (forcings and static attributes) for the time step $t$, $W$'s are the network weights, $b$'s are bias parameters, $y$ is the output to be compared to observations, $h$ is the hidden states, and $s$ is called the cell states of memory cells, which is unique to LSTM. Readers are referred to the literature for the detailed functionality of these units. Summarized briefly, $i$, $f$, $o$ control, respectively, when the input is significant enough to use, how long the past states should be remembered for, and how much the value in memory is used to compute the output. During training, $W$'s and $b$'s are adjusted using back-propagation through time (BPTT). In BPTT, the network is first unrolled over a prescribed length before the difference between the output and target propagates into the network. We used the LSTM implemented in Recurrent Neural Network library for Torch7 \citep{Leonard2015}, which is a scientific computing framework for the programming language Lua. We employed Dropout regularization, which randomly sets a fraction (dropout rate, $dr$) of its operand to $0$. Dropout prevents the co-adaptation of neurons and thus reduces overfitting. We used dropout regularization to non-recurrent links as in \cite{Zaremba2015}, a constant dropout mask to recurrent connections as in \cite{Gal2015}. We also implemented dropout for the memory cell as described in \cite{Semeniuta2016}. \\

At each time step, the network outputs one scalar value ($y^{(t)}$), which is compared to SMAP L3 passive product. The loss function to be minimized is the mean-squared error calculated for the time series: 
\begin{equation} 
L = \frac{1}{\rho}\sum\limits_{t=1}^\rho\mathbf{1}_{o}(t)[y^{(t)} - y^{*(t)}]^2
\end{equation} 
where $\mathbf{1}_{o}(t)$ is $1$ when time step $t$ has SMAP observation and is 0 otherwise, $\rho$ is the length of the time series, and $y^{*(t)}$ is SMAP observation. For computational efficiency and stability reasons, the training is done through "mini-batches": for each batch, a number of instances, or SMAP pixels, are randomly collected from the training set. The loss function is then averaged over all the instances in a batch. 




\subsection{Tests, Conventional Algorithms, and Evaluation Metrics}
In our temporal generalization test, the training set is SMAP data from April 2015 to March 2016. For computational efficiency, we picked 1 pixel from every 4 x 4 patch, resulting in a 1/16 coverage of CONUS. The test set is SMAP data for the same pixels, but for the period from April 2016 to March 2017. In the regular spatial generalization test, the training set is the same as described above, but the test set is the neighboring cells for the same period. In the regional holdout test, we trained models over 4 of the 18 2-digit Hydrologic Cataloging Units (HUC2s) and tested on others. This test challenges the ability of different methods to generalize across characteristically different climates and physiographic conditions. There are a large number of such 4-HUC2 combinations. As an initial investigation, we chose 4 of such combinations (C1-C4). Two combinations have a broad coverage of the range of Noah's bias over CONUS, while the other two cover only part of that range. These tests inform us the effect of biased training sets on generalization.


LSTM predictions are compared to three conventional methods: the least absolute shrinkage and selection operator (lasso), auto-regressive model (AR), and a single-layer feedforward Neural Network (NN), given the same inputs. Lasso, shorthanded as LR here, is multiple linear regression with a regularization term that penalizes large regression coefficients \citep{Tibshirani1994}. NN can construct nonlinear transformations of inputs, but does not have memory, and therefore cannot retain time dependencies. LR and NN are operated in (1) the CONUS-scale mode, where a single model is trained for the entire training set; and (2) point-by-point mode, with subscript "$_{p}$", where a separate model is trained for each pixel. AR models are trained only point-by-point (AR$_p$). More details are provided in SI Text S1. Three statistical metrics, bias (the time-averaged difference), root-mean-squared error (RMSE) and Pearson's correlation ($R$) are calculated between between predicted and SMAP-observed soil moisture on training and test sets separately. $R$ measures the agreement between simulated and observed climatology.

While short-term forecast employs observations to continuously update solution, long-term hindcast has no observations to use. As a "proof-of-concept" test of LSTM's appropriateness for long-term hindcast, we trained LSTM and AR$_p$ using 2 years of Noah-simulated soil moisture as the target, to hindcast to 10 years back.



\section{Results}
\subsection{Overall test performance}
\begin{figure}[h]
    \begin{minipage}{0.5\textwidth}
        \centering
        \includegraphics[width=1\linewidth]{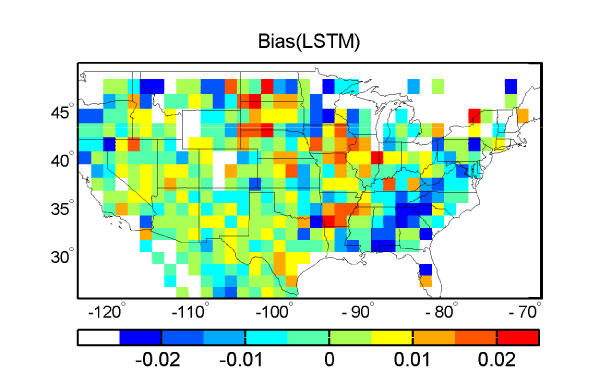}
        (a)
    \end{minipage}    
    \begin{minipage}{0.5\textwidth}
        \centering
        \includegraphics[width=1\linewidth]{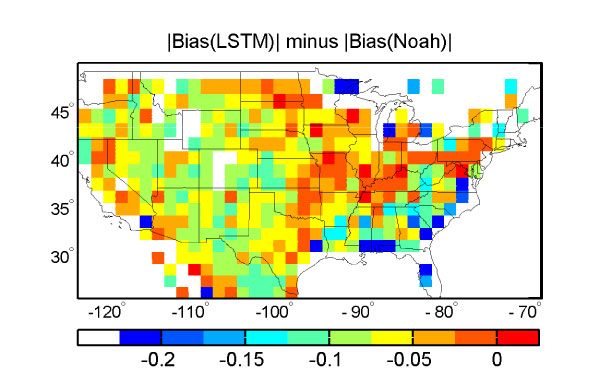}
        (b)
    \end{minipage}    
    \begin{minipage}{0.5\textwidth}
        \centering
        \includegraphics[width=1\linewidth]{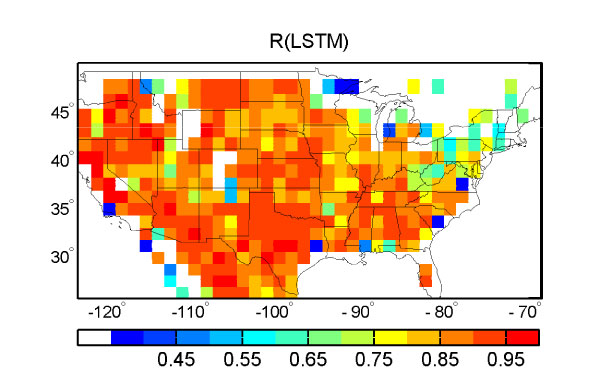}
        (c)
    \end{minipage}    
    \begin{minipage}{0.5\textwidth}
        \centering
        \includegraphics[width=1\linewidth]{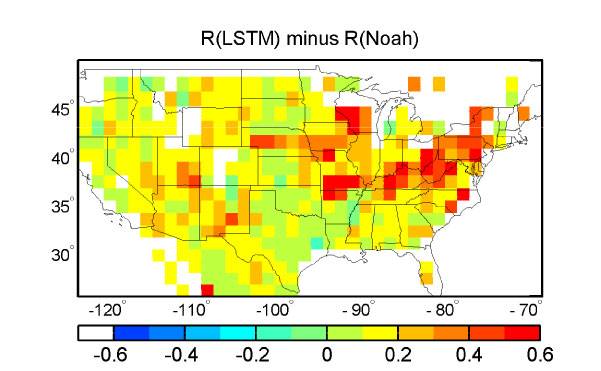}
        (d)
    \end{minipage}    
    \begin{minipage}{0.5\textwidth}
        \centering
        \includegraphics[width=1\linewidth]{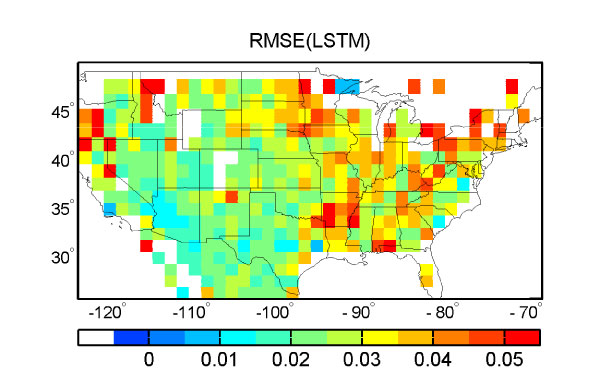}
        (e)
    \end{minipage}    
    \begin{minipage}{0.5\textwidth}
        \centering
        \includegraphics[width=1\linewidth]{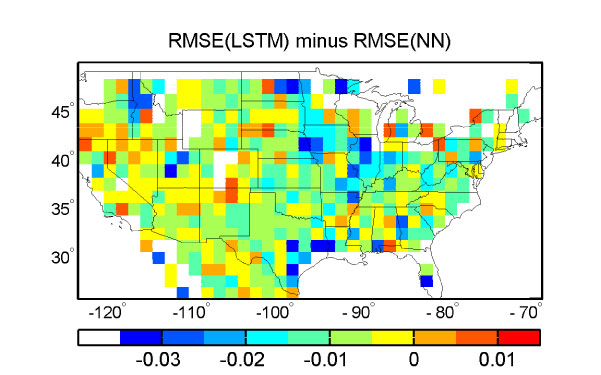}
        (f)
    \end{minipage}
\caption{Performance of LSTM predictions in the test set of the temporal generalization test. All metrics are evaluated against SMAP. (a) $bias(LSTM)=\overline{LSTM - SMAP}$ . Each pixel in this figure is patch of 4x4 SMAP pixels. Bias in most parts of CONUS is between -0.02 and 0.015; (b) Change of absolute bias due to LSTM correction. LSTM reduces the absolute bias significantly over CONUS; (c) LSTM anomaly correlation ($R(LSTM)$); (d) change of $R$ due to LSTM correction; (e) $RMSE(LSTM)$; (f) Since the RMSE improvement over Noah looks similar to panel b, here we show the difference in RMSE between LSTM and NN predictions. Maps of Noah's performance is provided in Figure S3 in SI.}
\label{fig_rmseMap}
\end{figure}

For the temporal generalization test, we note substantial improvement with respect to both bias and $R$ compared to Noah (Figure \ref{fig_rmseMap}). We report results from directly using the 0-10 cm Noah layer, although other choices are similar, as will be discussed later. The Noah solutions have a significant, spatially-varying bias in many parts of CONUS, as shown in Figure S3a in SI, especially in southeastern coastal plains (annotated in Figure S2 in SI). The LSTM correction reduces the bias by an order of magnitude, and mostly removed the spatial pattern of bias (Figure \ref{fig_rmseMap}a). We note there is a CONUS-scale spatial trend of larger reduction of absolute bias in the Eastern CONUS, except the southeast coast (Figure \ref{fig_rmseMap}b). The gradient appears to be related to the CONUS annual precipitation map, as it corrects Noah's bias to over-estimate in the arid west while over-estimate in the humid east (Figure S3a, also noted in \citep{Xia2015}).


LSTM does not only reduce bias but more noticeably improve the climatology, according to $R$ (Figure \ref{fig_rmseMap}c,d), which only concerns the comparison in temporal fluctuations. LSTM $R$ is mostly above $0.8$ and 50\% pixels are over 0.9, significantly above Noah. In most CONUS the $R$ improvement is greater than {\color{red}$0.1$}, while it can be $0.3\sim0.5$ in the Eastern half CONUS, especially the agricultural regions in central lowland and on the Appalachian Plateau (Figure \ref{fig_rmseMap}d). We note this map is no longer similar to the bias map of Noah, suggesting mechanisms that correct seasonality are different from those correcting bias. We hypothesize LSTM significantly improves soil moisture dynamics in agricultural regions, {\it e.g.}, irrigation, and the influence of shallow soils on the Appalachian highlands \citep{Fang2017}. On the other hand, over the majority of CONUS, the RMSE of LSTM is lower than 0.035 (Figure \ref{fig_rmseMap}e). A continental-scale west-to-east increasing trend in RMSE(LSTM) is apparent. The higher errors in the East may result from higher annual precipitation, which results in (i) higher annual-mean soil moisture, and (ii) high VWC, which reduces SMAP data quality. However, the Southeast regions facing the Atlantic has a rather low RMSE(LSTM). Figure \ref{fig_rmseMap}f suggests the improvement of LSTM over the one-layer NN is obvious, especially in the central lowland region and coastal plains. 


\begin{figure}[h]
\includegraphics[width=1\linewidth]{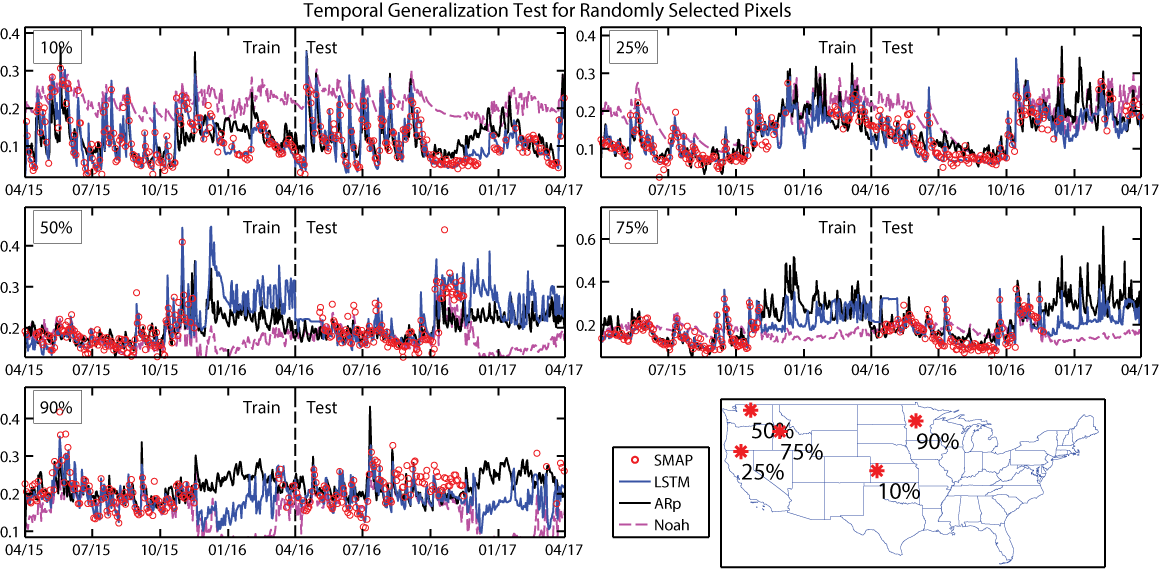}
\caption{Comparisons between SMAP observations and soil moisture predicted by LSTM, Noah, and AR$_p$ at 5 locations. We chose sites around 10-th, 25-th, 50-th, 75-th and 90-th percentiles as ranked by $R(LSTM)$. } 
\label{fig_lstm_ts}
\end{figure}

\begin{figure}[h]
\includegraphics[width=1\linewidth]{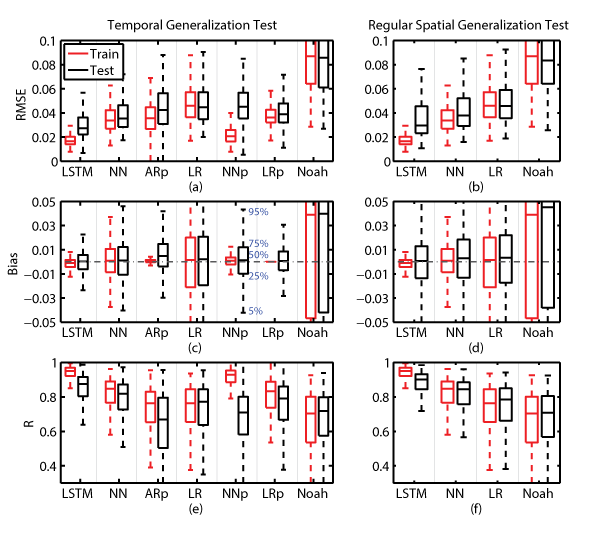}
\caption{Boxplots comparing LSTM, Noah, NN, AR, LR, $NN_{p}$ and $LR_{p}$ in the temporal generalization test and the regular spatial generalization test. Each box and whisker element summarizes SMAP pixels over CONUS with percentiles annotated in the panel. Y-axis limits truncate Noah boxes to focus on the central part of other boxes. The left column is the temporal generalization test, and the right column is the regular spatial generalization test. The three rows are for RMSE, Bias and $R$, respectively.} 
\label{fig_boxplots}
\end{figure}

\begin{figure}[h]
\includegraphics[width=1\linewidth]{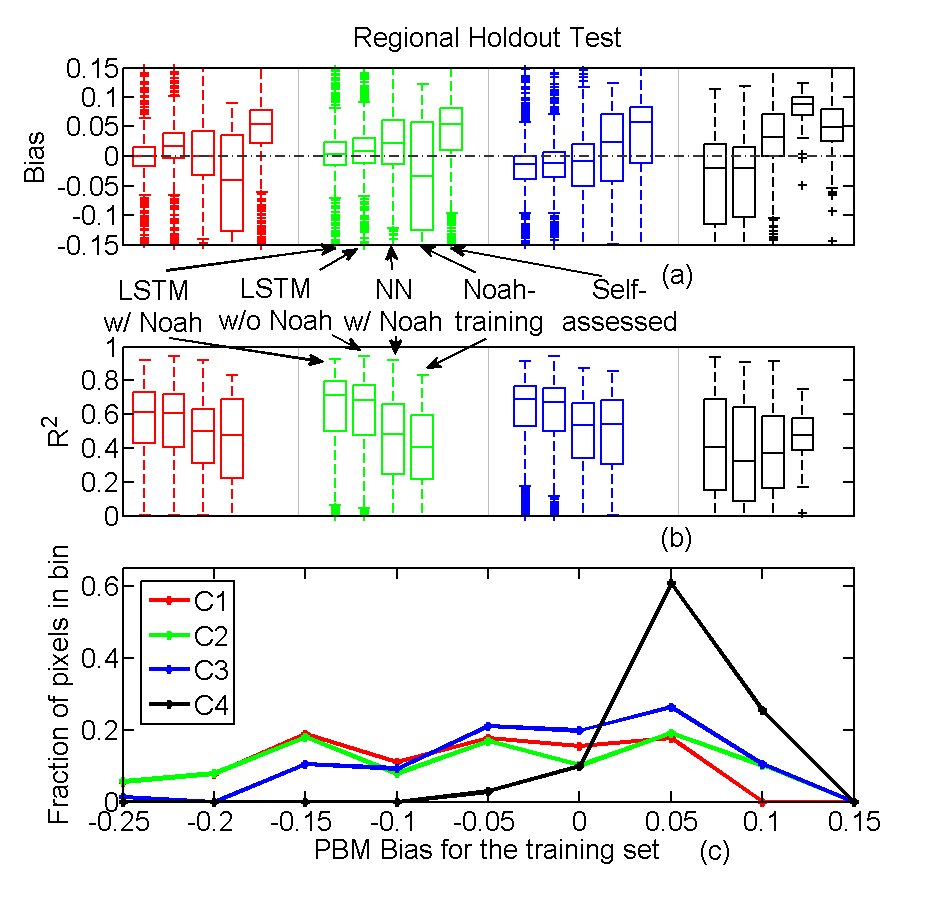}
\caption{(a) Test biases for cases C1-C4 (distinguished by different colors), which are four combinations of 4 HUC2s. Each group consists of 5 boxes, which are, respectively, LSTM with Noah among inputs, LSTM without Noah, NN with Noah in inputs, Noah in the training set, and self-assessed test bias. (b) test coefficient of determination ($R^2$). We note significantly better seasonality captured by LSTM in cases C1-C3, but not necessarily in C4; (c) Distributions of Noah's biases in the training sets for C1-C4.}
\label{fig_region}
\end{figure}

\subsection{Comparison of generalization capability with other methods}

In the temporal generalization test, time series prolonged by LSTM compares favorably against AR$_p$ across a wide range of LSTM performance levels $R(LSTM)$ (Figure \ref{fig_lstm_ts}). For the 10-th to even 75-th percentile pixel, LSTM is able to closely follow SMAP, except that peaks are under-estimated in the 50-th percentile pixel in 2016. The frequent rain events in April-May 2016 in Figure \ref{fig_lstm_ts}a and their recessions are well captured. For the 75-th percentile pixel, all peaks are captured, but we notice some over-estimation near the troughs in August 2016. In contrast, while AR$_p$ is also behavioral, we notice it often noticeably under-estimates the troughs, over-estimates the seasonal rising limbs and overshoots some peaks. In the 10-th percentile cell, AR$_p$ performs poorly between October 2016 and early 2017.

Summarized over CONUS, LSTM shows the lowest test RMSE and bias, and the highest $R$ (Figure \ref{fig_boxplots}), followed by NN, LR$_p$, AR$_p$, LR, NN$_p$ and lastly Noah. Neither the vertical interpolation procedure nor the choice of LSM (MOS or Noah) has much impact on LSTM's prolongation performance (see Figures S4 and S5 in SI). The test RMSEs of LSTM are 0.022, 0.027, 0.036 and 0.057 for the $25^{th}$, $50^{th}$, $75^{th}$ and $90^{th}$ percentile pixel, respectively (Figure \ref{fig_boxplots}a). With lasso regularization, LR has similar training and test RMSEs, but its 25-th percentile test RMSE is similar to the $75$-th percentile of LSTM's. Therefore, the more complex relationships permitted by LSTM are beneficial. The LR$_p$ improves from LR as it specializes in each pixel, and the lasso regularization appears to have prevented overfitting, but its error is still larger than the CONUS-scale LSTM. NN$_p$ and AR$_p$ appear more overfitted than LR$_p$. LSTM's test bias is only moderately smaller than that of NN, AR$_p$ and LR$_p$, but $R$ is much higher. 75\% and 50\% of $R$(LSTM) are greater than 0.80 and 0.87, respectively.  

Note AR$_p$ has sub-par performance in both training and test periods. The test RMSE box for AR$_p$ is wider, suggesting its formulation works well for some pixels but not so well in others. Furthermore, the extended proof-of-concept long-term hindcast experiment shows a similar contrast. LSTM has robust prolongation performance at a 10-year hindcast scale while AR$_p$ generates larger errors. Errors for both methods are independent of hindcast lengths, i.e., 10-year-prior hindcast error is not much different from 2-year hindcast (Text S2 and Figure S6 in SI). Meanwhile, in the regular spatial generalization test, LSTM again exhibits the smallest RMSE and bias (Figure \ref{fig_boxplots}c-d). The contrast in bias is smaller than the temporal test, but the $R$ comparisons are similar.

In 4-HUC2 combinations 1 and 2 (C1 and C2), the Noah bias covers a wide range from -0.25 to 0.15 cm$^3$/cm$^3$, which appears to be the whole range of the Noah biases we see over CONUS. In both cases, LSTM is able to greatly reduce the bias and improve soil moisture climatology (much higher $R$) compared to both NN and Noah (Figure \ref{fig_region}a-b). The boxes corresponding to LSTM bias are very narrow, and its centers are nearly 0. For the case C3, we note that it has few points with bias <-0.2, so for this HUC2 combination, the training set under-samples the Noah errors that lead to strong negative biases. As a result, LSTM's bias is no longer near 0, although still much better than NN's and Noah's. On the other hand, for C4, the training set is strongly biased. It lacks any basin with a Noah bias of <-0.1. We note the narrow box corresponding to Noah's bias in the training set for C4. Unsurprisingly, LSTM's performance deteriorates: LSTM is no longer able to correct the bias, and its range of bias is large. NN, similarly, also fails to correct the bias. We obtained LSTM's self-assessment of Noah bias by subtracting Noah's solution from LSTM's prediction. The self-assessed bias (Figure \ref{fig_region}a) has a range of bias which has little overlap with the Noah bias in the training set. This may be a signal we can utilize in the future to identify biased training sample and large predictive uncertainty.


\section{Discussion}

In many parts of CONUS, LSTM's RMSE is smaller than SMAP's design measurement accuracy. It appears even with 1 year of data over CONUS, when grouped together, has enough information to train an LSTM to hindcast SMAP data. A factor that contributes to such performance is the short memory length of soil moisture, which was found to range between 5 to 40 days \citep{Orth2012} in previous work. As a result, two years of data, when grouped together, contain many complete wet-dry cycles. The hindcast quality should improve as SMAP data increases. On a side note, because true random noise cannot be predicted in the test set, it follows that SMAP L3's RMSE could be below 0.027 in 50\% of CONUS. Also, the official SMAP data quality flag labels the forested Southeast Coastal Plains and South Atlantic regions as "not recommended" quality (Figure S3). Our LSTM has a RMSE of 0.02-0.035 there, which suggest SMAP may be functional in these regions, but the impacts of the retrieval algorithm should be carefully examined.

It seems non-recurrent NN can already remove a large part of bias by capturing how environmental factors lead to certain type of biases. However, NN cannot maintain time dependencies, which may explain its performance difference from LSTM. Therefore, we argue an advantage of LSTM originates from its recurrent nature. Meanwhile, alternative recurrent models, e.g., AR and moisture loss functions \citep{Koster2017}, are profoundly useful due to their interpretability, parsimony and great value in "nowcasting" or short-term forecasting (see \cite{Koster2017} for a solid application), but they require continuous updates by observations to avoid drift from true dynamics. At one-year scale, injected data already has little effects on hindcast solutions. For longer-term hindcast, pattern-based methods like LSTM appear to be more suitable.

Previous soil moisture comparisons mainly focused on anomalies, but the prevalent bias with Noah's surface moisture simulations can cause large errors in downstream users such as weather modeling \citep{Massey2016}. The continental-scale bias pattern suggests some systematic errors with Noah's model structure/parameters. Some hypotheses include (i) Noah's soil pedo-transfer functions are fundamentally inadequate in resolving regionally heterogeneous soil responses to rainfall, which could explain the need for calibration in most large-scale flood forecasting systems; or (ii) groundwater flow, which is important in thick-soiled, high-infiltrating capacity regions like the southeast \citep{Fang2016}, is not properly simulated in LSMs \citep{Clark2015}. However, LSTM appears to be able to integrate information from raw data and compensate for the inadequate representation uniformly over CONUS.


Conventional statistical wisdom suggests that simpler models are more robust and models with high degrees of freedom may be easily overfitted. However, the present work shows the CONUS-scale deep learning networks have smaller test errors than three alternative methods trained point-by-point. In fact, we hypothesize an important strength of LSTM originates from its flexibility to simultaneously learn from a large and heterogeneous collection of data and identify commonalities and differences. Its generalization capability stems from building internal models (in the attribute space) to capture biases and temporal fluctuations. For our regional holdout test, creating these internal models does not seem to require having all combinations of climates and physical attributes in the training set, as the HUC2s have distinct climates, topography, landcover and soils.

\section{Conclusion}
We have trained a CONUS-scale LSTM network to predict SMAP data. This network is capable of correcting spatially-heterogeneous model bias as well as climatological errors between Noah-simulated and SMAP-observed top-surface soil moisture, creating a CONUS-scale seamless moisture product that has high fidelity to SMAP. Despite having high degrees of freedom, when properly regularized, LSTM exhibits better generalization capability, both in space and time, than linear regression, auto-regressive models, and a one-layer neural network. Its test error approaches the instrument accuracy limit with SMAP. LSTM will be helpful in long-range soil moisture hindcasting or forecasting, weather modeling, and data assimilation. Its generalization capability arises from building internal models from physical attributes and synthesis of climate forcing. It does not necessarily require similar examples in the training set. Unless the training set is strongly biased, LSTM has a good chance of success.

\section{Limitations and Future Work}
As a first paper using LSTM in hydrology, this work is by no means a thorough investigation. Optimization is certainly possible. Our work does not address the question about the accuracy of SMAP data, which is addressed by other studies. The hindcast performance with respect to capturing soil moisture during drought should be further examined with {\it{in-situ}} data. We should further assess LSTM's performance in comparison with regionally-trained simpler models. The implications of low LSTM RMSEs in forested region warrants further investigations.

\section{Acknowledgment}
Data for SMAP can be downloaded from SMAP's data repository. Work is supported by seed grants from Penn State College of Engineering and Institute for CyberScience. Shen is partially supported by Office of Biological and Environmental Research of the US Department of Energy under contract DE-SC0010620. We thanks NVIDIA Corporation for donating a Tesla K40 GPU for this research.

\begin{acronyms}
\acro{SMAP}
Soil Moisture Active Passive
\acro{DL}
Deep learning
\acro{LSTM}
Long Short-Term Memory
\acro{LR}
Linear Regression
\acro{ANN}
Artificial Neural Network
\end{acronyms}

\bibliography{reference}

\renewcommand{\thefigure}{S\arabic{figure}}
\renewcommand{\thetable}{S\arabic{table}}
\setcounter{figure}{0}

\journalname{arXiv}


\supportinginfo{Prolongation of SMAP to Spatio-temporally Seamless Coverage of Continental US Using a Deep Learning Neural Network}

\authors{Kuai Fang\affil{1}, Chaopeng Shen\affil{1}, Daniel Kifer\affil{2}, Xiao Yang\affil{2}}

\affiliation{1}{Department of Civil and Environmental Engineering,Pennsylvania State University, University Park, Pennsylvania, USA.}
\affiliation{2}{Department of Computer Science and Engineering, Pennsylvania State University, University Park, Pennsylvania, USA.}

\correspondingauthor{Chaopeng Shen}{cshen@engr.psu.edu}

\setcounter{equation}{0}
\section*{Contents}
\begin{enumerate}
\item Text S1. Technical details about conventional statistical methods. 
\item Figure S1. Comparison between the structures of Long Short-Term Memory (LSTM) and simple recurrent neural network (RNN). 
\item Figure S2. Map of SMAP's data quality flag,  with annotations for geographic regions on the continental US
\item Figure S3. Noah bias and RMSE evaluated against SMAP over CONUS.
\item Figure S4. Performance of training using Noah soil moisture interpolated to 5 cm depth.
\item Figure S5. Comparing LSTM models created with Noah or MOS models as inputs.
\item Text S2. Proof-of-concept test for the potential of LSTM for long-term hindcast. 
\item Figure S6. Performance of LSTM and AR for the synthetic long-term hindcast experiment
\item Table S1. Predictors employed by LSTM, lasso-regularized linear regression and one-layer feedforward neural network
\end{enumerate}


\newpage

\setlength{\abovedisplayskip}{4pt}
\setlength{\belowdisplayskip}{4pt}
\section*{Text S1. Technical Details about Conventional Methods}
We compared the Long-Short Term Memory (LSTM) network to the least absolute shrinkage and selection operator (lasso), auto-regressive moving average model (AR), and a single-layer feedforward Neural Network (NN), given the same inputs. 
Because lasso is essentially a regularized linear regression, it is shorthanded as LR in our paper. The equation for estimating the parameters for LR is:
\begin{equation}
    \beta^{LR}=\operatornamewithlimits{argmin}\limits_{\beta^{LR}_0,\beta^{LR}}{\Big({\frac{1}{2N}\sum\limits_{i=1}^{N}(\theta^o_i-\beta^{LR}_0-x_i^T\beta^{LR})^2+\lambda\sum\limits_{j=1}^n \left | {\beta^{LR}} \right | \Big)}}
\end{equation}
where $\theta^o$ is the SMAP soil moisture product, $\beta^{LR}$ are coefficients for the LR model, $\lambda$ is a regularization parameter that determines how much penalty is applied on large coefficients, and $x$ contains exogenous inputs including temperature, precipitation, wind, downward shortwave and long wave radiations, specific humidity, and Noah-simulated potential evapotranspiration, evaporation, and runoff. In alternative models that we examined, we also tested removing the list of Noah outputs. The regularization parameter ($\lambda$) is determined experimentally to minimize the test error, and a value of 0.002 is found to be appropriate for LR, and point-by-point LR (LR$_p$). 

We have added point-by-point auto-regressive model with exogenous inputs into the comparisons, meaning a separate model is trained for each SMAP pixel. We did not consider moving average models because our focus is on the potential of the method for long-term forecast, while moving-average models require observations to calculate residuals. The equation for the AR is:
\begin{align}
    \theta_t = c + \epsilon_t + \sum\limits_{i=1}^p {\alpha_i\theta_{t-i}} + \sum\limits_{k=1}^r {\gamma_kx_{k,t}} 
\end{align}
where $c$ is a constant, $t$ is the time step, $\theta$'s are soil moisture observations, $p$ is the order of the auto-regression, $\alpha$ and $\gamma$ are coefficients that will be estimated for each SMAP pixel, and $x_t$ are $r$ forcing inputs as indicated above. For our long-term hindcast test. We could include static attributes in this equation but since they are static in time they will be absorbed by the constant $c$, and because we are training point-by-point there is no reason to consider them. During parameter estimation (training) stage, observations are used to update the past states ($\theta_{t-i}$). In the long-term hindcast (testing) stage, because there is no observation, $\theta$ are the AR-predicted values. The model has to recursively apply the forecast equation to proceed in time. We varied $p$ from 0 to 5 and identified the value that gave the smallest testing error for each site.

The one-layer feedforward neural network (NN) is simply a linear combination of inputs $x$ and a transformation:
\begin{align}
    \theta^{NN}(t)=f(W_{NN}x+b)
\end{align}
where $W_{NN}$ is the weights of the neural network, b is a constant coeffient and $f$ is a nonlinear transformation, in this case tan-sigmoid ($tansig$). We regularized NN using early stopping and L2-norm regularization. A regularization parameter of 0.002 was found to be give the smallest test root-mean-squared error (RMSE). NN and its point-by-point version, NN$_p$, have a linear hidden layer of size 100 and 30, respectively, as larger hidden size results in more over-fitting.


\begin{figure}[H]
\includegraphics[width=1\linewidth]{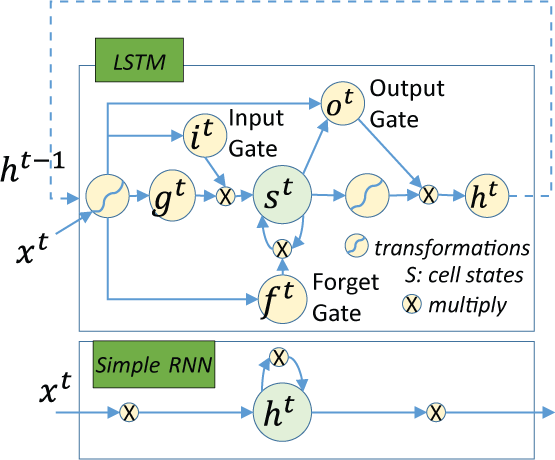}
\caption{Comparing an Long Short-Term Memory (LSTM) unit with simple recurrent neural network (RNN). The transformations from inputs to $i$, $f$, $o$ are sigmoidal functions. From inputs to $g$ and from $s$ to $h$ the transformation is $tanh$. $\otimes$ means multiplication by weights. Main point: the conventional design of RNN only iteratively update the hidden state. The design of gates in LSTM allows it to learn when to forget past states, and when to output, thus addressing the issue of slow training of front node with RNN. Figure is modified from [{\it Greff et al.}, 2015]. } 
\label{fig_sup_Prcp}
\end{figure}

\begin{figure}[H]
\includegraphics[width=1\linewidth]{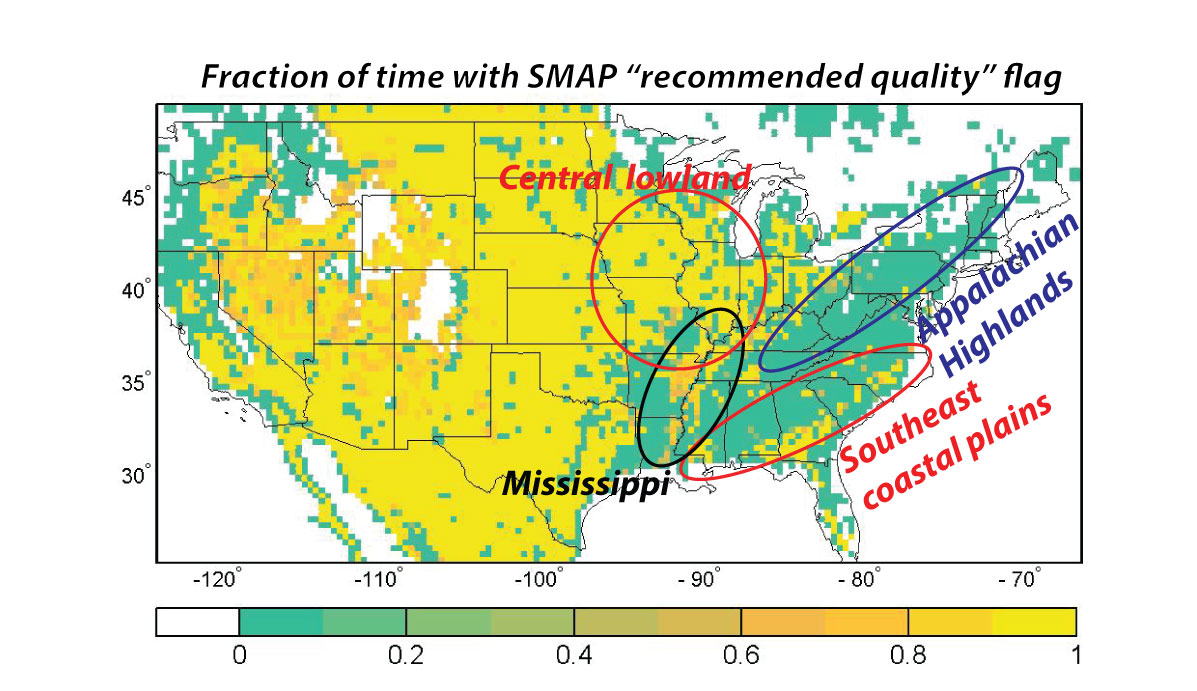}
\caption{SMAP data quality with geographic regions annotated on the map for reference. The values shown is the time-averaged SMAP "recommended quality" flag. We notice that Applachian Highlands and Southeast Coastal Plains are both mostly flagged as having bad quality, but LSTM's root-mean-squared-error from SMAP L3, RMSE(LSTM) are in the range of 0.02-0.035 in the South Appalachian and Coastal Plains according to Figure 1. Because random error cannot be captured in the test, it suggests SMAP quality may be not as bad as thought. However, this finding and the potential influence from models in the retrieval algorithm need to be thoroughly evaluated.} 
\label{fig_sup_qualFlag}
\end{figure}

\begin{figure}[H]
    \begin{minipage}{0.5\textwidth}
        \centering
        \includegraphics[width=1\linewidth]{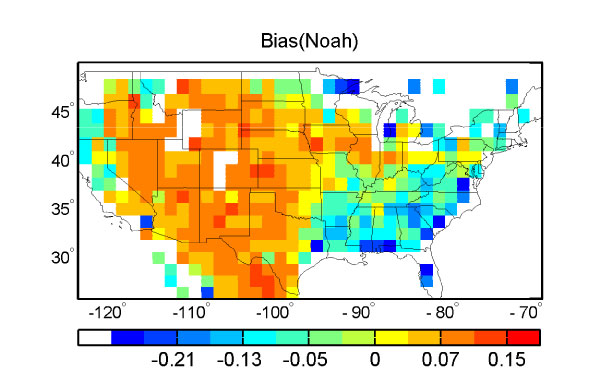}
        (a)
    \end{minipage}    
    \begin{minipage}{0.5\textwidth}
        \centering
        \includegraphics[width=1\linewidth]{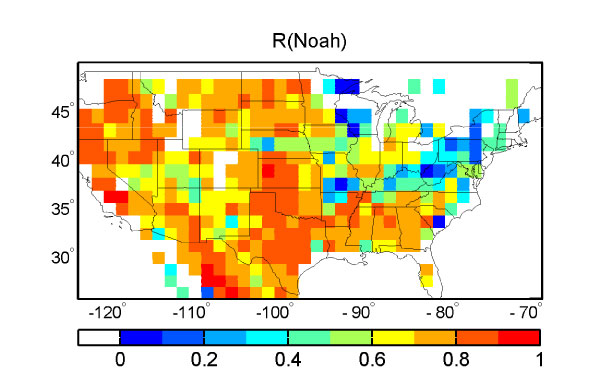}
        (b)
    \end{minipage}    
\caption{Performance of Noah evaluated against SMAP in the testing set of the temporal generalization test. (a) Bias: the time-averaged value of Noah-predicted soil moisture, interpolated to 5 cm, and SMAP L3 product; (b) anomaly correlation coefficient between Noah-predicted soil moisture and SMAP L3 product.}
\label{fig_NOAHMap}
\end{figure}

\begin{figure}[H]
\includegraphics[width=1\linewidth]{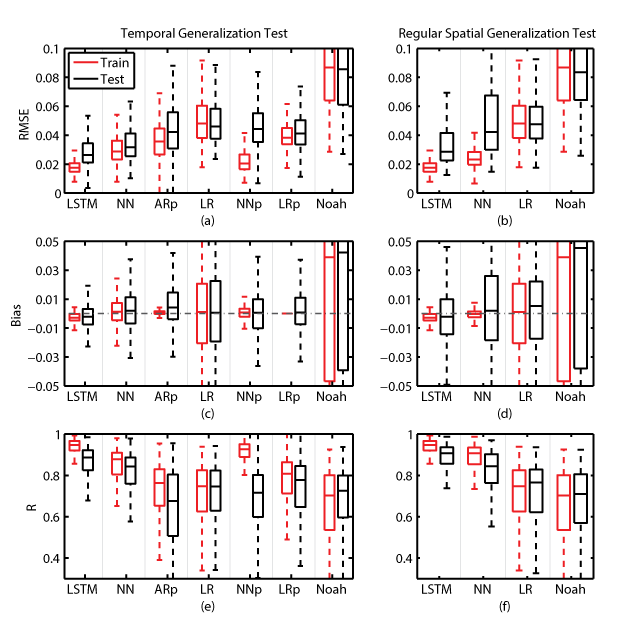}
\caption{Same as Figure 3 but the Noah-simulated values are linearly interpolated to 0-5 cm. We tested several interpolation methods: (i) directly using 0-10 cm data; (ii) linear (2-point) and cubic vertical interpolation (3-point) using top layers; and (iii) integral interpolation: we determined a $2^{nd}$-order (or $3^{rd}$) polynomial whose integral in these layers agree with Noah-simulated values. This Figure shows method (i), whose results are very similar to those reported in Figure 3, while other interpolation methods also generate similar results}
\label{fig_boxplots2}
\end{figure}

\begin{figure}[H]
\includegraphics[width=1\linewidth]{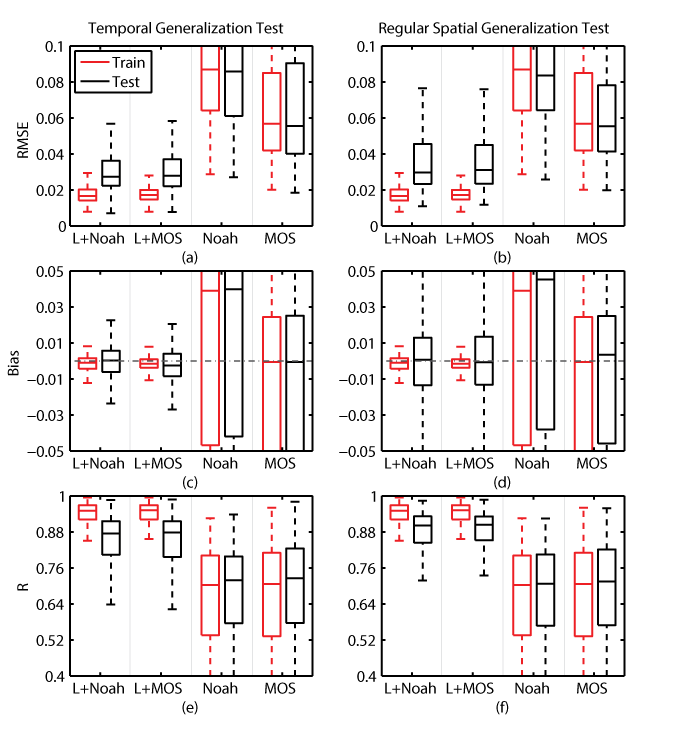}
\caption{Comparisons between LSTM models with Noah (L+Noah) and MOS (L+MOS) solutions as inputs. Both Noah and MOS solutions are obtained from North American Land Data Assimilation System (NLDAS). The distinction between "train" and "test" for Noah and MOS only means the different time periods for which the metrics are calculated. Noah and MOS have comparable performance in simulating moisture climatology. It appears MOS generally has smaller root-mean-squared error (RMSE) and smaller bias. However, using which model in the inputs does not seem to have a noticeable impact on the test performance of the LSTM models.}
\end{figure}

\newpage

\section*{Text S2. Proof-of-concept test for the potential of LSTM for long-term hindcast}
Since SMAP has a limited time span, we conducted a proof-of-concept experiment that examines the potential of LSTM for multi-year-scale soil moisture hindcasting and compare it to point-by-point auto-regressive models (AR$_p$). These synthetic experiments are not thorough in performance optimization, as true hindcasting will involve auxiliary satellite-based observations and {\it{in-situ}} data. Both LSTM and AR$_p$ are trained in 2015-2016 with Noah-simulated soil moisture as the target and climate forcing as the inputs. Based on the temporal generalization test described in the main text, we removed all Noah-simulated fields from the inputs, and, since AR$_p$ does not require any static attributes like topography and soil texture, we also removed such attributes from LSTM's inputs. We added two types of synthetic noise to Noah solutions: a Gaussian white noise (with standard deviation = 0.04) and a relative error. Neither types of noise is auto-correlated. The formulae for the relative error is:
  \begin{alignat}
\theta_s=\theta_{Noah}*(1+\epsilon)
\end{alignat}
where $\theta_s$ is the synthetic observation to be treated as the learning target, $\theta_{Noah}$ is the top 10 cm soil moisture simulated by Noah, and $\epsilon\sim N(0,0.07)$ is a Gaussian relative error term. 

\begin{figure}[H]
\includegraphics[width=1\linewidth]{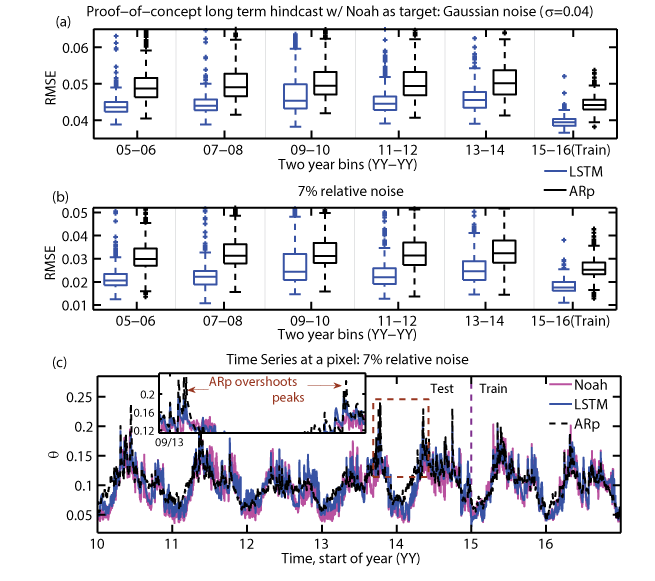}
\caption{Proof-of-concept long-term hindcast tests with Noah-simulated soil moisture as the target. (a) boxplot comparing errors (evaluated against Noah) for the 10-year hindcast. Noah solution is contaminated by a Gaussian noise with a standard deviation ($\sigma$) of 0.04. The RMSEs are calculated for each 2-year period and are grouped over CONUS to form the boxplot. Note that error does not increase as hindcast length increases, i.e., during 2005-2006, the errors are not greater than those in 2013-2014; (b) same as (a) but for a $7\%$ relative noise; (c) time series for Noah, LSTM and AR$_p$ at a pixel. We only show 5 years of hindcast for clarity of the plot. The zoomed-in panel on (c) (corresponding to the brown box in the main plot) highlights how AR$_p$ over-estimates the two soil moisture peaks. Meanwhile, AR$_p$ seems to have dampened small-scale fluctuations}
\label{fig_longterm}
\end{figure}

The results show that with two years of training data, LSTM can well learn the soil moisture dynamics of Noah (Figure \ref{fig_longterm}a-b). The median error for the white-noise case only slightly increases from ~0.04 in the training period, which is almost equal to the added noise) to 0.043 in 2005-2006. Importantly, the hindcast noise does not increase as a function of hindcast length, {\it{i.e.}}, distance from the first synthetic observation. The AR$_p$ also works decently, with a median error around 0.049 in 2005-2006. Its error is also not influenced by hindcast length, perhaps because soil moisture dynamics simulated by Noah has only limited memory length. However, LSTM is still noticeably stronger as 85-th percentile of LSTM's error in 2005-2006 is less than 25-th percentile error of AR$_p$. The LSTM boxes are much narrower than those of AR$_p$. Also, note that we only created one LSTM model for the continental U.S. (CONUS). In addition, LSTM can make use of static attributes to differentiate between locations with different soil textures and land covers, but AR$_p$ cannot. Therefore, the performance of LSTM may further improve as these attributes are included. LSTM may compensate for the lack of attributes by summarizing information from climate forcings, as climate features co-vary with physical attributes. Figure \ref{fig_longterm}c compares the hindcast time series at a pixel. We note that AR$_p$ tends to over-predict major peaks but under-predict the rise limbs. LSTM well captures the troughs but AR$_p$ may over-predict the troughs. 

As Noah has simpler dynamics and less unknown variables than real systems, it is easier to learn so it is not surprising the errors are close to the added Gaussian noise. The larger error of AR$_p$ during the training period suggest its formulation is not flexible enough to completely reproduce the dynamics of Noah. These results shown here mainly illustrates that LSTM has a great potential for long-term hindcasting. It appears from our results that since soil moisture has short memory, hindcasting to one year is not very different from hindcasting to 10 years. However, the training data should adequately sample plausible soil moisture dynamics.

\begin{longtable}{p{0.1\textwidth} p{0.4\textwidth} | p{0.1\textwidth} p{0.4\textwidth}}
\caption{Predictors used in the training of LSTM, lasso-regularized linear regression, and one-layer feedforward neural network}
\label{tab:predictors}\\
\hline
	\multicolumn{4}{c}{NLDAS Model Outputs}  \\  \hline
	ALBDO & Albedo & RCSOL & Soil moisture parameter in canopy conductance \\  
	ARAIN & Liquid precipitation & RCT & Temperature parameter in canopy conductance \\  
	ASNOW & Frozen precipitation  & RSMACR & Relative soil moisture availability control factor \\  
	AVSFT & Average surface skin temperature & RSMIN & Minimal stomatal resistance \\  
	BGRUN & Subsurface runoff (baseflow) & SBSNO & Sublimation (evaporation from snow) \\  
	CCOND & Canopy conductance & SHTFL & Sensible heat flux \\  
	CNWAT & Plant canopy surface water  & SNOD & Snow depth \\  
	GFLUX & Ground heat flux  & SNOHF & Snow phase-change heat flux \\  
	LAI & Leaf area index & SNOM & Snow melt \\  
	LHTFL & Latent heat flux & SNOWC & Snow cover  \\  
	LSOIL & Liquid soil moisture content (non-frozen) & SOILM & Soil moisture content \\  
	MSTAV & Moisture availability  & SSRUN & Surface runoff (non-infiltrating) \\  
	NLWRS & Net longwave radiation flux (surface) & TRANS & Transpiration \\  
	NSWRS & Net shortwave radiation flux (surface) & TSOIL & Soil temperature \\  
	PEVPR & Potential latent heat flux & VEG & Vegetation \\  
	RCQ & Humidity parameter in canopy conductance & WEASD & Water equivalent of accumulated snow depth \\ 
	RCS & Solar parameter in canopy conductance & \  & \  \\  
	\hline
	\hline 
	\multicolumn{4}{c}{NLDAS Forcing}  \\  \hline
	ACOND & Aerodynamic conductance  & EVP & Evaporation \\  
	ACPCP & Convective precipitation hourly total & HGT & Geopotential height \\  
	APCP & Precipitation hourly total & PEVAP & Potential evaporation hourly total \\  
	CAPE & above ground Convective Available Potential Energy & PRES & Surface pressure \\  
	CONVfrac & Fraction of total precipitation that is convective & SPFH & Specific humidity \\  
	DLWRF & Longwave radiation flux downwards & TMP & Temperature \\  
	DSWRF & Shortwave radiation flux downwards  & UGRD & Zonal wind speed \\  
	\hline
	\hline 
	\multicolumn{4}{c}{SMAP Flags}  \\  \hline
	albedo & Albedo & vegewater & Vegetation water content \\  
	coast & Coastal proximity & roughness & Roughness \\  
	waterbody & Radar water body fraction & staWater & Static water \\  
	landcover & Landcover classes & urban & Urban area \\  
	mount & Mountainous terrain & vegetation & Dense vegetation \\  
	\hline
	\hline 
	\multicolumn{4}{c}{Geographic attributes}  \\  \hline
	Bulk & Bulk density & Irri & Irrigation \\  
	Capa & Soil capacity & Sand & Sand fraction \\  
	Clay & Clay fraction & Silt & Silt fraction \\  
	LULC & NLCD 2001 land cover and use type & & \\  
	\hline
\end{longtable}


\end{document}